\documentclass[twoside,11pt]{article}

\usepackage[utf8]{inputenc}

\usepackage[frozencache=true,cachedir=.]{minted}
\usepackage{caption}
\usepackage{subcaption}

\usepackage{indentfirst}
\usepackage{lmodern}

\usepackage[x11names,dvipsnames,table]{xcolor} 

\usepackage{etoolbox}
\makeatletter
\AtBeginEnvironment{minted}{\dontdofcolorbox}
\def\dontdofcolorbox{\renewcommand\fcolorbox[4][]{##4}}
\makeatother

\RequirePackage{etex}
\usepackage{float}
\usepackage{paralist}
\usepackage{amsmath}
\usepackage{enumitem}
\usepackage{pgfplots}

\usepackage{tikz,tkz-euclide}
\usepackage{tikz-3dplot}
\usetikzlibrary{matrix}
\usetikzlibrary{shapes.multipart,calc}
\usetikzlibrary{decorations.markings}
\usetikzlibrary{decorations,arrows,shapes,shapes.misc}
\usetikzlibrary{decorations.pathreplacing}

\newcommand{\R}{\mathbb{R}}
\newcommand{\M}{\mathrm{M}}
\newcommand{\N}{\mathrm{N}}
\newcommand{\D}{\mathrm{D}}

\DeclareUnicodeCharacter{2207}{\nablachar}

\newcommand*\new{\color{black}}

\definecolor{om}{RGB}{0,0,200}

\definecolor{com}{RGB}{200,0,0}

\newcommand{\zp}[2][p]{%
  \!\IfStrEqCase{#1}{%
    {p}{\left(}%
    {c}{\:\left[}%
    {a}{\left\{}%
    {abs}{\left|}%
    {n}{\left\|}%
    {s}{\left<}%
    {i}{\left[\!\left[}%
  }[\left(]%
    #2%
  \IfStrEqCase{#1}{%
    {p}{\right)}%
    {c}{\right]}%
    {a}{\right\}}%
    {abs}{\right|}%
    {n}{\right\|}%
    {s}{\right>}%
    {i}{\right]\!\right]}%
  }[\right)]%
} 
\newcommand{\keops}{\texttt{KeOps}}
\newcommand{\cpp}{\texttt{C++}}
\newcommand{\cuda}{\texttt{CUDA}}
\newcommand{\cppcuda}{\texttt{C++/CUDA}}
\newcommand{\pytorch}{\texttt{PyTorch}}
\newcommand{\matlab}{\texttt{Matlab}}
\newcommand{\numpy}{\texttt{Numpy}}
\newcommand{\gnur}{\texttt{GNU R}}
\newcommand{\python}{\texttt{Python}}
\newcommand{\tvm}{\texttt{TVM}}

\usepackage[abbrvbib]{jmlr2e}
\renewcommand{\cite}[2][]{\citep[#1]{#2}}

\pgfplotsset{compat=1.15} 

\firstpageno{1}

\usepackage{lastpage}
\jmlrheading{}{}{}{}{}{}{Benjamin Charlier, Jean Feydy, Joan Alexis Glaunès, François-David Collin and Ghislain Durif}
\ShortHeadings{Kernel Operations on GPU, without memory overflows}{Charlier, Feydy, Glaunès, Collin and Durif}

\begin{document}

\title{Kernel Operations on the GPU, with Autodiff,\\ without Memory Overflows}

\author{\name Benjamin Charlier* \email benjamin.charlier@umontpellier.fr \\
       \addr IMAG\\
       Université de Montpellier, CNRS\\
       Montpellier, France
       \AND
       \name Jean Feydy* \email jean.feydy@ens.fr \\
       \addr DMA\\
       École Normale Supérieure\\
       Paris, France
       \AND
       \name Joan Alexis Glaunès* \email alexis.glaunes@parisdescartes.fr \\
       \addr MAP5 \\
       Université de Paris, CNRS \\
       Paris, France\\
       \name François-David Collin \email francois-david.collin@umontpellier.fr\\
       \name Ghislain Durif \email ghislain.durif@umontpellier.fr\\
       \addr IMAG\\
       Université de Montpellier, CNRS\\
       Montpellier, France \hfill
       \textnormal{* equal contribution}
       }
       
\editor{}

\maketitle

\begin{abstract}
The \keops{} library provides a fast and memory-efficient GPU support for tensors whose entries are given by a mathematical formula, such as kernel and distance matrices.
\keops{} alleviates the {\new main} bottleneck of tensor-centric libraries 
for kernel and geometric applications: memory consumption. 
It also supports automatic differentiation and outperforms standard GPU baselines,
including \pytorch{} \cuda{} tensors or the \texttt{Halide} and \tvm{} libraries. 
\keops{} combines optimized \cppcuda{} schemes with binders for high-level languages: \python{} (\numpy{} and \pytorch), \matlab{} and \gnur{}. As a result, high-level ``quadratic'' codes can now scale up to large data sets with millions of samples processed in seconds. 
\keops{} brings graphics-like performances for kernel methods and is  freely available on standard repositories (PyPi, CRAN). To showcase its versatility, we provide tutorials in a wide range of settings online at 
\href{https://www.kernel-operations.io}{\texttt{www.kernel-operations.io}}.
\end{abstract}



\section{Introduction}
\label{sec:intro}

Recent advances in machine learning have been driven by the diffusion of two pieces of software: automatic differentiation and GPU backends for tensor computations. Today, thanks to {\new e.g.} the \texttt{TensorFlow} or \texttt{PyTorch} libraries \cite{tensorflow, pytorch}, users routinely perform gradient descent on functions that involve millions of parameters.

{\new
These high-level \texttt{Python} frameworks unlock the
use of massively parallel hardware for machine learning research.
Under the hood, they rely on \texttt{C++} routines that are
often supported by hardware manufacturers:
the \texttt{cuBLAS} and \texttt{cuDNN} libraries edited by Nvidia
provide the binaries for linear algebra
and convolutions that power a majority of deep learning models.
In practice, the presence of a complete software stack
(from low-level binaries to well-documented \texttt{Python} libraries)
is a prerequisite for the widespread adoption
of a research idea by the machine learning community.
The \texttt{KeOps} library intends to provide such a
solid numerical foundation
for all methods that involve large distance or kernel matrices.
A motivating example is the computation of pair-wise interactions
of the form:
\begin{align}
    a_i ~\gets~ 
    \textstyle\sum_{j=1}^\N \exp( - \|x_i-y_j\|^2 / 2\sigma^2)~ b_j 
    ~=~\textstyle\sum_{j=1}^\N k(x_i,y_j) \,b_j 
    \qquad \text{for} \quad i=1,\dots,\M
    \label{eq:gaussian_product}
\end{align}
where $\M$ and $\N$ range from a hundred to a billion,
$x_1$, \dots, $x_\M$ and $y_1$, \dots, $y_\N$ all belong
to a common vector space $\R^\D$,
the signals $b_1$, \dots, $b_\N$ are real numbers 
and $k(x,y)$ is a Gaussian kernel of deviation $\sigma > 0$. 
This operation is often understood as a matrix-vector
product with a Gaussian kernel matrix $(K_{i,j}) = (k(x_i,y_j))$, or
equivalently as a convolution with the kernel function $k$
that is sampled on the point clouds $(x_i)$ and $(y_j)$.

To perform this computation, 
common practice
is to create an explicit $\M$-by-$\N$ buffer $(K_{i,j})$
and compute Eq.~(\ref{eq:gaussian_product})
with a linear algebra routine.
Unfortunately, this method requires the
\emph{storage} of the kernel matrix 
as a contiguous array in memory
and does not scale when $\M$ and $\N$ are
in the order of $50$k or more.
To work around this problem,
a common strategy is thus to decompose Eq.~(\ref{eq:gaussian_product})
as a collection of smaller matrix-vector products
using a \texttt{Python} or \texttt{Matlab} ``\texttt{for}'' loop.
This method alleviates memory issues but remains inefficient
in spaces of dimension $\D \leqslant 100$:
the \emph{transfer} of the tiles of the kernel
matrix $(K_{i,j})$ between different layers of GPU memory
remains a narrow bottleneck for computations.
}
\new{
\keops{} leverages efficient \texttt{C++} schemes
from the graphics literature \cite[Chapter 31]{GpuGems3}
to streamline the use of registers and reach
optimal performance in this setting.}
It is fast, transparent to use and targets a single yet powerful abstraction: semi-symbolic arrays whose entries {\new $M_{i,j}$ are given by a mathematical formula
``$F(x_i, y_j)$'', evaluated on data arrays that are indexed by line and column numbers ``$i$`` and ``$j$''.}
It can be called from the major scripting languages used in the scientific community: \python{} (\numpy{} and \pytorch), \matlab{}, and \gnur{}. In this introduction, we focus on the \python{} interface and
refer to our documentation for additional tutorials,
applications and benchmarks.

\section{\texttt{\textbf{KeOps}} Purpose and Usage}
\label{sec:keops_usage}

\emph{A generic reduction framework.} 
The workhorse of the \keops{} library is 
a {\new \texttt{C++}} engine for generic reductions
{\new on sampled data}. 
Let us assume that we have at hand:

\medskip

\begin{compactenum}\setlength\itemsep{.1cm}
\item 

{\new

\textbf{parameters}: a collection $p^1\in\R^{d_{p}^1}, \dots, p^P\in\R^{d_{p}^P}$ of vectors;
\item \textbf{$\boldsymbol i$-variables}: a collection $x^1\in\R^{M\times d_{x}^1}, \dots, x^X\in\R^{M\times d_{x}^X}$ of matrices, with rows indexed by $i\in\zp[i]{1, \M}$ (hence for each $k$ and $i$, $x^k_i$ is a vector in $\R^{d_{x}^k}$);

\item \textbf{$\boldsymbol j$-variables}: a collection $y^1\in\R^{N\times d_{y}^1}, \dots, y^Y\in\R^{N\times d_{y}^Y}$ of matrices, with rows indexed by $j\in\zp[i]{1, \N}$ (hence for each $k$ and $j$, $y^k_j$ is a vector in $\R^{d_{y}^k}$);
\item a vector-valued \textbf{symbolic formula} $F(p^1, \dots, p^P, x^1_i, \dots, x^{X}_i, y^1_j, \dots, y^Y_j)\in\R^{d_{\text{out}}}$;
\item a \textbf{reduction} operation such as a sum, max, argmin, log-sum-exp, etc.

}

\end{compactenum}


\noindent
Then, a single call to the {\new \texttt{KeOps C++}} engine allows users to evaluate the expression:
\begin{equation}
    a_i ~\gets~ \operatorname{Reduction}_{j=1,\dots,\N}\limits 
    \big[ F(p^1, \dots, p^P, x^1_i, \dots, x^{X}_i, y^1_j, \dots, y^Y_j)  \big] 
    \quad  \text{for} \quad i=1,\dots,\M  \label{eq:keops_genred}
\end{equation}
efficiently, with a linear memory footprint on GPUs and CPUs.
As illustrated in our gallery of tutorials, 
this level of generality allows \keops{} to handle off-grid convolutions,
$k$-nearest neighbors classification, 
$k$-means clustering and many other tasks.



\emph{The \texttt{LazyTensor} abstraction.}
The ``\texttt{LazyTensor}'' wrapper for \texttt{NumPy} arrays
and \texttt{PyTorch} tensors
{\new lets users} specify computations
along the lines of Eq.~(\ref{eq:keops_genred})
with a tensor-like interface. 
For instance, we can specify the
Gaussian matrix-vector product of Eq.~(\ref{eq:gaussian_product}) with:

\begin{minted}[bgcolor=blue!4,linenos,mathescape,tabsize=4,firstnumber=1,stripnl=false,fontsize=\small]{python}
from pykeops.torch import LazyTensor  # Wrapper for PyTorch Tensors
x_i  = LazyTensor(x[:,None,:])      # (M,D) Tensor -> (M,1,D) Symbolic Tensor
y_j  = LazyTensor(y[None,:,:])      # (N,D) Tensor -> (1,N,D) Symbolic Tensor
D_ij = ((x_i - y_j)**2).sum(dim=2)  # (M,N,1) Symbolic matrix of squared distances
K_ij = (- D_ij / (2 * s**2)).exp()  # (M,N,1) Symbolic Gaussian kernel matrix
a    = K_ij @ b  # Genuine torch Tensor. (M,N,1) @ (N,D) = (M,D)
\end{minted}

In the script above, no computation is performed at lines 4 and 5:
lazily, the \keops{} engine builds up a symbolic formula $F$
encoded as a string attribute of the \texttt{LazyTensor} \texttt{K\_ij}. 
{\new The lazy chain is only terminated by
the virtual matrix product of line 6, a generic reduction
that triggers the real computation:}
no intermediate $\N$-by-$\M$ buffer is created in the global device memory. 

Note that variable types ($i$-, $j$-variable or parameter) 
are inferred from the shapes of the input tensors at lines~2 and~3:
{\new in practice, symbolic tensors are
as easy to use as sparse matrices.
The \texttt{LazyTensor} wrapper turns a dense array into a symbolic matrix whose
axes -3 and -2 are understood as ``virtual'' dimensions;
a reduction on these axes is the signal that triggers
a call to the \texttt{KeOps C++} engine.
}


As showcased on our website and at the end of this paper, \keops{}
scripts for kernel and geometric applications generally outperform
their \numpy{} and \pytorch{} counterparts
by several orders of magnitude while keeping a linear memory footprint. 
\texttt{LazyTensors} support a wide range of mathematical operations that mimic the usual
interface for \texttt{NumPy} arrays and \texttt{PyTorch} tensors.
They fully support broadcasting and batch dimensions, 
as well as a decent collection 
of reduction operations: \texttt{.sum()}, 
\texttt{.logsumexp()}, \texttt{.max()} and \texttt{.min()} 
{\new  but
also \texttt{.argmin()} or \texttt{.argKmin(K=...)} that return
the \emph{indices} of the smallest (or K-smallest) elements
of the rows of a symbolic tensor.}

{\new 
\emph{Inner engine.}
Internally, \texttt{KeOps} creates an optimized \texttt{C++} code
for every new reduction and formula $F$ that it encounters.
Binaries are then compiled and stored on the hard drive for later use:
compilation relies on the standard \cuda{} stack (\texttt{nvcc}, \texttt{gcc} and/or \texttt{clang} compilers) and is only performed once per reduction.
}

\emph{Backpropagation.}
{\new 
Crucially, \keops{} supports automatic differentiation up to arbitrary orders of differentiation:
a new binary is created automatically for every new partial derivative
that is required by the user's computations. }
This mechanism is fully integrated with the
\texttt{torch.autograd} engine and lets users
``backprop'' through \keops{} calls using
the usual \texttt{torch.autograd.grad()}
and \texttt{.backward()} methods.

\newpage

\section{Performance evaluation}
\label{sec:performances}

{\new 
\texttt{KeOps} is geared towards computations
that fit the mould of Eq.~(\ref{eq:keops_genred}).
In this specific context, it combines
a fully transparent interface with state-of-the-art performance.
To illustrate this, we compare \keops{}
to similar scientific computing libraries
-- \texttt{PyTorch},
\texttt{TensorFlow}, \texttt{Halide} \cite{Halide} 
and \texttt{TVM} \cite{tvm}
-- on a simple benchmark:
the Gaussian kernel matrix-vector product of Eq.~(\ref{eq:gaussian_product}) with an increasing number of points $\M=\N$ in dimension $\mathrm{D}=3$. All experiments are performed with \texttt{float32} precision on a Nvidia RTX 2080 Ti GPU, with the exception of the PyTorch-TPU column that was run in Google Colab;
code is available on our repository in the
\href{https://github.com/getkeops/keops/tree/master/benchmarks}{\texttt{benchmarks}}
folder.
Our \href{https://www.kernel-operations.io/keops/_auto_benchmarks/index.html}{\texttt{gallery}} also includes comparisons
with JAX \cite{jax} and domain-specific libraries
such as FAISS \cite{faiss} for K-Nearest Neighbors search.
}
\vspace{.2cm}

\centerline{\small
\rowcolors{2}{black!10}{white}%
\renewcommand*{\arraystretch}{1.2}%
\begin{tabular}{ |c|cccccc|c| } 
    \hline
     & \texttt{PyTorch} & \texttt{PyTorch-TPU}
     & \texttt{TF-XLA} & \texttt{Halide}
     & \texttt{TVM}    & \textbf{\texttt{PyKeOps}}
     & \texttt{KeOps++} \\
    \hline
    $\N=10\text{k}~\,$ 
    & 9\,ms & 10\,ms
    & 13\,ms & 1.0\,ms
    & 3.80\,ms & \textbf{0.7\,ms}
    & 0.4\,ms \\ 
    $\N=100\text{k}$ 
    & \texttt{out of mem} & \texttt{out of mem}
    & 89\,ms & 34.1\,ms
    & 36.8\,ms & \textbf{15.0\,ms}
    & 14.6\,ms\\ 
    $\N=1\text{M}~\,\,$ 
    & \texttt{out of mem} & \texttt{out of mem}
    & \texttt{out of mem} & 3.8\,s
    & 2.79\,s & \textbf{1.39\,s}
    & 1.38\,s\\ 
    Lines of code 
    & 5 & 5
    & 5 & 15
    & 17 & 5
    & 55  \\ 
    \shortstack{Interface\\\vspace{.15cm}} &
    \shortstack{\vspace{.05cm}\\\numpy{}-\\like} & \shortstack{\numpy{}-\\like} &
    \shortstack{\numpy{}-\\like} & 
    \shortstack{\texttt{C++}\\\vspace{.15cm}} &
    \shortstack{low-level\\\python{}} & \shortstack{\textbf{\numpy{}}-\\\textbf{like}} &
    \shortstack{\texttt{C++}\\\vspace{.15cm}} \\
    \hline
\end{tabular}
}\vspace{.2cm}

{\new 
As evidenced by this table, \keops{} turns \texttt{NumPy}-like scripts
into highly competitive binaries.}
Going further, it can be neatly interfaced with the iterative linear solvers of the \texttt{Scipy} \cite{scipy} or \texttt{GPytorch} \cite{gpytorch} libraries
and supports the specification of cluster-wise block-sparsity patterns: this allows users to solve large kernel linear systems efficiently, with applications to geology (Kriging), imaging (splines), statistics (Gaussian processes) and data sciences (kernel methods). 

{\new 

\section{Intended Use, Limitations and Future Works}

\keops{} fills a specific but important niche in machine learning research.
Unlike most other compilers for deep learning computations, 
such as \texttt{Halide} and \texttt{TVM}, it is meant to be used directly
by theorists of the machine learning community.
Our focus on the simple yet powerful concept
of \emph{symbolic matrices} allows us to keep
a transparent interface, while being more efficient
than the generalist \texttt{PyTorch} and \texttt{XLA} frameworks
on a wide range of computations.
In future works, we intend to add support for approximation schemes
such as the Nyström and FFM methods \cite{nystrom,fastfuriousmethod},
beyond the block-sparse truncation rule
that is currently supported.
These features will be valuable to many researchers in the field,
while being out of scope for most deep learning libraries.

The core strength of the \texttt{KeOps} \texttt{C++} engine is that it
optimizes the use of GPU registers for computations
that fit the template of Eq.~(\ref{eq:keops_genred}).
In practice, it is therefore of limited use in situations
where a kernel matrix can fit in memory and 
is re-used a large number of times, 
or when the evaluation of the formula ``$F(p,x_i,y_j)$''
takes a significant amount of time.
Keeping in mind the motivating example of Eq.~(\ref{eq:gaussian_product}),
we believe that \texttt{KeOps} will be most useful
for computations that involve $10$k points or more
in a space of dimension $\D \leqslant 100$.
Going forward, our main priority
will be to ease the deployment of pre-compiled \keops{} binaries,
reduce compilation times to at most a handful of seconds per routine
and add support for the newly released \texttt{CUDA} Tensor cores.



}

\section*{Acknowledgements}

The three first authors are the project leaders: they contributed equally to the library
and its documentation. 
The authors also thank Alain Trouvé, whose theoretical work in shape analysis
was the first motivation for the development of the \keops{} engine.

\vskip 0.2in
\bibliography{biblio}

\end{document}